%% file: ijcai23.tex
\documentclass{article}

\input{macros}

\usepackage{ijcai23}

\usepackage{times}
\usepackage{soul}
\usepackage{url}
\usepackage[hidelinks]{hyperref}
\usepackage[utf8]{inputenc}
\usepackage[font={small}]{caption}
\usepackage{graphicx}
\usepackage{amsmath}
\usepackage{amsthm}
\usepackage{amssymb}
\usepackage{booktabs}
\usepackage{natbib}
\usepackage{algorithm}
\usepackage{algorithmic}
\usepackage[switch]{lineno}

\title{Planning Multiple Epidemic Interventions with Reinforcement Learning}

\author{
Anh Mai$^1$
\and
Nikunj Gupta$^2$\and
Azza Abouzied$^1$\And
Dennis Shasha$^2$
\affiliations
$^1$New York University Abu Dhabi\\
$^2$New York University
\emails
\{anh.mai, ng2531, azza\}@nyu.edu,
shasha@cims.nyu.edu
}

\begin{document}
\maketitle

\begin{abstract}

Combating an epidemic entails finding a plan that describes when and how to apply different interventions, such as mask-wearing mandates, vaccinations, school or workplace closures. An optimal plan will curb an epidemic with minimal loss of life, disease burden, and economic cost. Finding an optimal plan is an intractable computational problem in realistic settings. Policy-makers, however, would greatly benefit from tools that can efficiently search for plans that minimize disease and economic costs especially when considering multiple possible interventions over a continuous and complex action space given a continuous and equally complex state space. We formulate this problem as a Markov decision process. Our formulation is unique in its ability to represent multiple continuous interventions over any disease model defined by ordinary differential equations. We illustrate how to effectively apply state-of-the-art actor-critic reinforcement learning algorithms (PPO and SAC) to search for plans that minimize overall costs. We empirically evaluate the learning performance of these algorithms and compare their performance to hand-crafted baselines that mimic plans constructed by policy-makers. Our method outperforms baselines. Our work confirms the viability of a computational approach to support policy-makers.
\end{abstract}

\input{00-introduction}

\input{01-application}
\input{04-eval}
\input{07-related-works}

\input{06-conclusion}

\bibliographystyle{named}
\bibliography{ijcai23}

\appendix 
\input{08-appendix.tex}

\end{document}

%% file: macros.tex
\usepackage{longtable}
\usepackage{xspace}
\usepackage{xcolor}
\usepackage{enumitem}

\usepackage{amsmath}
\usepackage{amsfonts}

\usepackage{booktabs}
\usepackage{graphicx}
\usepackage{url}
\usepackage{caption,subcaption}

\usepackage{times}  
\usepackage{helvet}  
\usepackage{courier}  

\newcommand{\epi}{{\sc EpiPolicy}\xspace}

\newcommand{\sir}{{{\textsf{SIR}}}\xspace}
\newcommand{\sirv}{{{\textsf{SIRV}}}\xspace}
\newcommand{\cf}{{{\textsf{C15}}}\xspace}
\newcommand{\sira}{{{\textsf{SIR-A}}}\xspace}
\newcommand{\sirva}{{{\textsf{SIRV-A}}}\xspace}
\newcommand{\cfa}{{{\textsf{C15-A}}}\xspace}
\newcommand{\sirb}{{{\textsf{SIR-B}}}\xspace}
\newcommand{\sirvb}{{{\textsf{SIRV-B}}}\xspace}
\newcommand{\cfb}{{{\textsf{C15-B}}}\xspace}

\definecolor{demoyes}{rgb}{0, 0.4, 0}
\definecolor{demono}{rgb}{0, 0.4, 0}
\newcommand{\demoyes}{{\color{demoyes}{\Large$\bullet$}}}
\newcommand{\demono}{-}

%% file: 00-introduction.tex
\section{Introduction}

Public health policymakers face a myriad of challenges when designing and implementing intervention plans to curb the spread of an epidemic. First, each disease is unique: what works to contain an Ebola outbreak --- a disease spread through direct contact with infected bodily fluids --- is different from what works to contain a flu outbreak. Second, the efficacy and effectiveness of interventions may vary widely even for similar diseases: the flu vaccine's effectiveness can vary from 40\% to 60\% depending on how well the vaccines are matched to the current season's flu strains \cite{flu}. Third, the disease or our understanding of its dynamics may evolve as new disease variants arise. The new variants may entail changes in intervention plans to reflect the changes in the disease's transmissibility or severity \cite{covid-variant}. Finally, quantifying the cost and benefit of a combination of interventions is difficult
even after an extensive post-hoc analysis~\cite{eco}
In short, there are no \textit{template} intervention plans that can be universally and directly applied across all disease outbreaks, regions and populations. 

Nevertheless, in practice, research supporting policy makers often consists of the simulation of a small set of predefined and relatively coarse intervention plans on a carefully-calibrated epidemiological model to assess and compare the economic cost and disease burden of each plan \cite{feg}. While this approach has the virtue of simplicity, 
it disregards the large space of potential policies. 

\begin{figure}[t]
    \centering
    \includegraphics[width=1\linewidth]{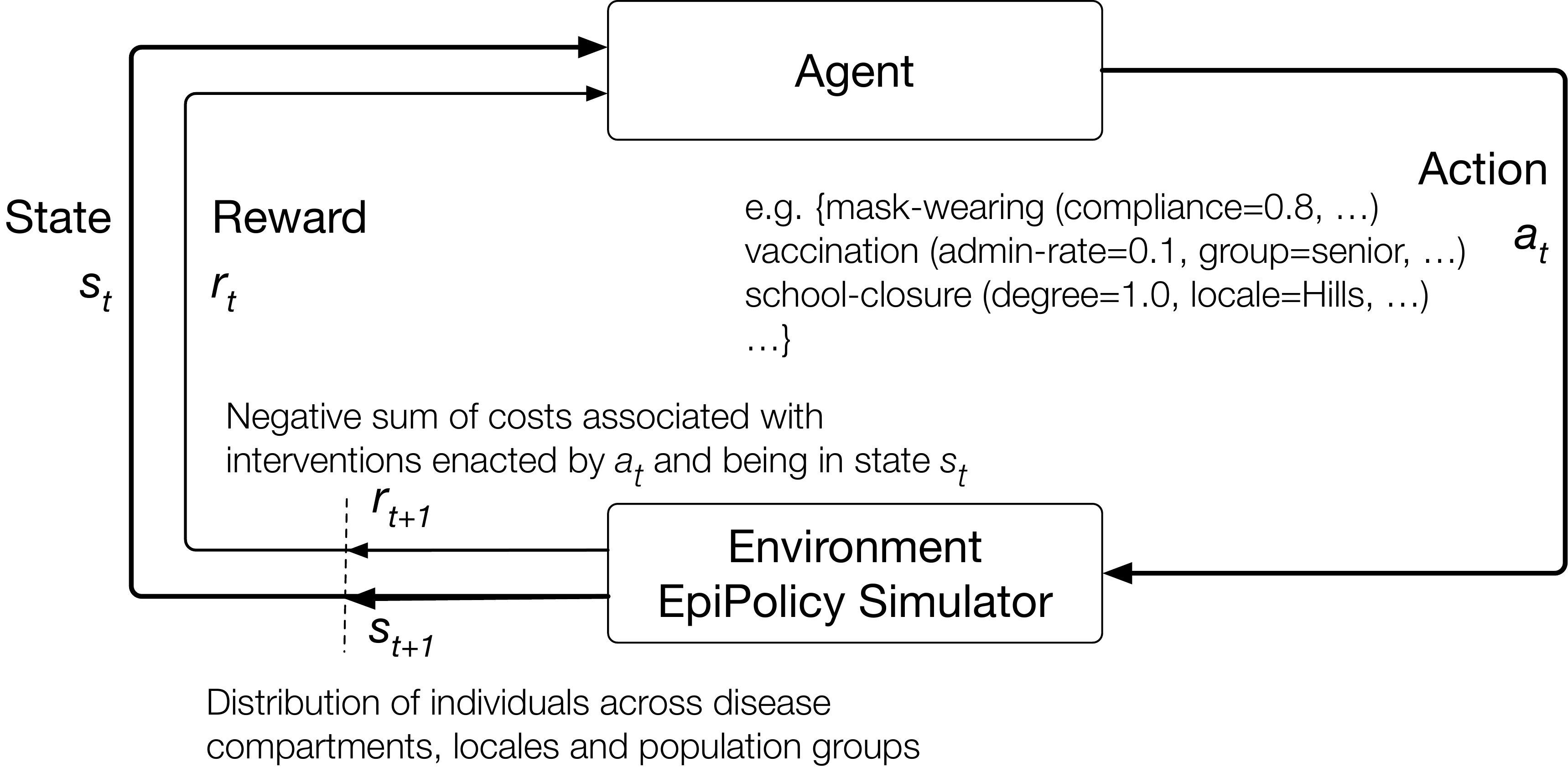}
    \caption{The agent-environment interaction in the Markov decision process formalizing the epidemic planning problem.}
    \label{fig:mdp}
\end{figure}

In contrast to the approach of choosing a plan from a set of predefined ones, we consider the following algorithmic challenge: \textit{ can we automatically and efficiently search for an optimal schedule of interventions that minimizes overall disease burden and economic cost?} An exact solution to this combinatorial search problem is intractable. With a one-year planning horizon of week-long timesteps, and only three binary interventions (e.g. close or open schools; close or open borders; enforce or relax mask-wearing mandates), there are $8^{52}$ plans to consider! Furthermore, many interventions have inherently continuous parameters (e.g. the number of vaccines to administer daily ranges from 0 to the available number of doses, or the distance between individuals in a physical distancing intervention may range from 0 to 10 meters) and policy-makers may disagree on how to discretize them.

Theoretical approaches that rely on dynamic programming techniques \cite{dp} or Pontryagin's maximum principle \cite{Perkins, Lemecha} neither scale nor generalize to  complex disease models (i.e. a large continuous state space) with a diverse set of candidate interventions. Recent works have examined the application of reinforcement learning (RL) to intervention planning. We can classify these works into ones that built disease simulators that enable control by an RL agent~\cite{Varun, Pieter}, or
 
programmable optimization toolboxes~\cite{EpOp}.
None of these works allows a large continuous disease state space, a multi-intervention continuous action space, and the ability to simulate different diseases and interventions as shown in Table \ref{tab:related_works}.

By contrast, we rely on an existing framework, \epi ~\cite{epipolicy}, that can simulate (i) disease-spreading interactions across regions for a given disease model and population and (ii) programmatically-defined interventions with free parameters that determine their effects and costs. This allows us to empirically examine the rich problem of controlling \textbf{multiple} interventions with RL that none of the prior works have examined.

\paragraph{Contributions \& Paper Outline.}
We present two main contributions:

\noindent \textit{(i)~~ A demonstration of how to use reinforcement learning to construct epidemic intervention plans over complex continuousdisease and population models, with multiple interventions}. To achieve this, we  formulate the problem as a Markov Decision Process (MDP) in Section \ref{sec:mdp}. We explore the space of possible RL approches in Section \ref{sec:approaches}, to select two algorithms (PPO, SAC) that fit our problem and we empirically tune them (Section \ref{sec:experiment}).  We present promising empirical results on a wide range of epidemic planning problems of varying complexity both in terms of the state space and action space (Section \ref{sec:results}).

\noindent \textit{(ii)~~ A benchmark for reinforcement learning and epidemiology researchers (Section \ref{sec:benchmark}).} The environments in our benchmark represent real disease models and interventions. Moreover, the benchmark can be easily extended to include other disease models or interventions. The code, data, and experiments can be found in our GitHub repository\footnote{\url{https://github.com/huda-lab/RL-Epidemic-Benchmark}}. The benchmark was motivated by and built in consultation with public health officials, so it should be useful to computational epidemiology researchers. In addition, the benchmark  provides a testing point for RL algorithms allowing further research into improving their performance.

%% file: 01-application.tex
\section{Problem Formulation}
\label{sec:mdp}

We frame the  epidemic plan optimization problem as a Markov Decision Process (MDP),
where an \textit{agent} learns 
an (approximately) optimal \textit{policy} or plan by interacting with a simulator of 
the natural disease environment. Figure \ref{fig:mdp} illustrates the agent-environment interaction in our application. The MDP formalizes the sequential decision-making process of epidemic planning, where an action --- a full parameterization of the set of interventions --- at each time step (e.g., a week)  influences the immediate costs  (e.g., the costs associated with vaccinating a certain number of individuals)  and rewards (e.g.,the avoidance of the cost of sick days due to increases in the infectious populations). Thus, each action influences both  the subsequent state and  future costs and rewards. 

\subsection{The Markov Decision Process}
An MDP is a 4-tuple $\langle\mathcal{S}, \mathcal{A}, p(s_{t+1} | s_t, a_t), r(s_t, a_t, s_{t+1}) \rangle$:


\noindent\textbf{$\blacktriangleright$ State:} $\mathcal{S}$ is the state space. In our application, a state $s_t$ is the distribution of the population across different disease compartments (e.g., infected, recovered, hospitalized, etc.) including their regional (e.g, a particular state or province) and group subdivisions (e.g., adult, senior, child, male, female, etc.).
    
\noindent\textbf{$\blacktriangleright$ Action:} $\mathcal{A}$ is the action space. An action $a_t$ is the set of applied interventions at time $t$ and their parameter values (e.g. 82\% school closure).
    
\noindent\textbf{$\blacktriangleright$ State-transition:} $p(s_{t+1} | s_t, a_t)$ is a function $p: \mathcal{S} \times \mathcal{S} \times \mathcal{A} \rightarrow [0,1]$. The probabilities given by $p$ characterize the dynamics of the environment. In our setting, this function is evaluated by executing a deterministic simulator with the state $s_t$ and action $a_t$ as inputs for a single time step. fixed current state and action $s_t, a_t$. 
    Since the future state $s_{t+1}$ depends only on the current state $s_t$ and the applied action $a_t$, the Markov property is satisfied.

\noindent \textbf{$\blacktriangleright$ Reward:} $r(s_t, a_t, s_{t+1})$ is a function $r: \mathcal{S} \times \mathcal{A} \times \mathcal{S} \rightarrow \mathbb{R}$. The reward of applying action $a_t$ to state $s_t$ to reach state $s_{t+1}$ is simply the negative of the sum of the costs associated with implementing the interventions in the action $a_t$ and the costs associated with being in a certain state $s_t$ to state $s_{t+1}$ for the time step.

\subsubsection{The Simulator}
\label{sec:env}

 An epidemic simulator for compartmental disease models approximates the progression of the disease in the face of the interventions. 
 Concretely, it defines the behavior of the state-transition function, $p$, of the MDP.

 For our application, we require a simulator that:

\noindent (i)~~ uses a deterministic compartmental disease model. Typical compartments will be susceptible (S), infected (I), and recovered (R), with known transition rates such as recovery rate and infection rate. The compartmental model describes how disease spreads in a population with predefined demographic and geographic characteristics using ordinary differential equations (ODEs).

\noindent (ii)~~simulates the results of programmatically-defined and parameterized interventions (the actions) in terms of their effects on the transition rates and on the distribution of the population across the compartments. The behavior (effects and costs) of an intervention should be modifiable by its \textit{control parameters}. 

\noindent (iii)~~exposes the simulator's internal state in terms of the  distribution of the population across the different compartments, its incurred costs as a function of the cost of each state (e.g., the estimated dollar cost of $x$ individuals being sick), and the cost of interventions applied (e.g., each vaccine administered costs $y$ dollars). 


The first requirement follows from scalability  considerations. 
Stochastic  population models better capture randomness but their differential equations are more expensive to solve. In addition, they are not as scalable as deterministic models for RL training where many experiences need to be simulated. Agent-based models are also difficult to scale and are often limited to population sizes of at most one million.

\begin{figure}
\centering
\includegraphics[width=1\linewidth]{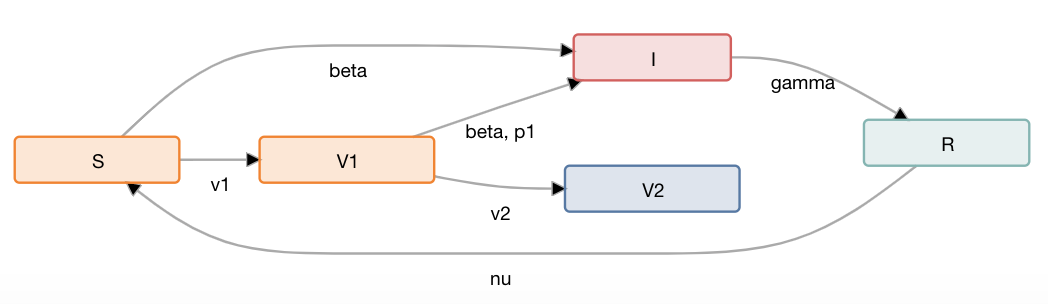}
\caption{The compartmental SIRV model with five compartments: susceptible (S), infectious (I), recovered (R) and two vaccinated compartments (V1, V2) for a disease with a two-dose vaccination regiment. Note the model has a transition edge from R to S annotated with the  immunity loss transition rate $\nu$ indicating the possibility of reinfection in this model.}
\label{fig:sirv}
\end{figure}

Figure \ref{fig:sirv} illustrates a compartmental disease model with four compartments: susceptible (S), infectious (I), recovered (R) and two vaccinated compartments (V1, V2) one for each dose in a disease with a two-dose vaccination regiment. Transition rates are the ODE parameters that control the progression of the population from one compartment to another. For example, three parameters in Figure \ref{fig:sirv}: $\beta$  the transmission rate, $\nu$  the loss of immunity rate, and $v_1$  the first-dose vaccination rate, together control the rate of change in the susceptible (S) compartment through the following ODE: $\frac{dS}{dt} = \frac{-\beta IS}{N} + \nu R - v_1 S$. Here $N$ is the total population count and $S, I, R$ are the number of individuals in their respective compartments at a given time. 

We assume that the transition rates are known and provided a priori. Inferring their values, sometimes called model calibration, is its own sub-field within epidemiology \cite{cali}.

The second requirement enables the RL agent to explore and control \textit{how} each intervention is applied by setting its control parameters. 

Finally, the third requirement (exposure to a simulator's internal state) enables the RL agent to evaluate the rewards of an intervention plan.

We use \epi as our epidemic simulator because it satisfies the above requirements~\cite{epipolicy, xrds}.

We note that in our formulation both \textbf{\textit{the action space and the state space are continuous}} and bounded. For example, interventions like school closures are controlled by a degree of the school closure parameter that ranges from 0 to 1. Given a fixed population size $N$, the number of individuals across the compartments and their regional or demographic subdivisions (e.g. male/female or child/adult/senior) range from $0$ to $N$\footnote{ODEs may result in real, non-integer, individual counts within compartments.}

\noindent\textbf{$\blacktriangleright$ Plan or Schedule. } An \textit{epidemic intervention plan} is simply the sequence of actions $a_0, ... a_T$ applied over a fixed time horizon $T$. Given an initial state $s_0$, the MDP, and the plan enacted by the agent, we can construct a trajectory: $s_0, a_0, r_0, s_1, a_1, r_1, ... s_T, a_T, r_{T}$. The cumulative reward of the plan is simply the (discounted) sum of all the rewards from $r_0$ to $r_{T}$: $r_0 + r_1 \gamma + r_2 \gamma^2 + ... + r_T \gamma^T$ where $\gamma \in [0,1]$ is the discount factor. In epidemic modelling, an action can have significant long-term consequences (like uncontrolled disease spread). For that reason, we set the discount factor at 0.99 to reflect the importance of mitigating the costs of an epidemic not only in the short term but also toward the end of the time horizon. 

\subsection{The Agent \& Solution Strategies}
\label{sec:approaches}

The goal of the \textit{agent} is to find an optimal epidemic intervention plan that minimizes overall cost. While formulating  epidemic planning as an MDP problem is straightforward, solving the MDP problem is not. We found that only Actor-Critic methods efficiently solve MDPs for planning epidemic interventions: analytical approaches do not extend beyond simple epidemiological models and tree search approaches suffer from inefficient sampling which require significant computational resources.

\paragraph{Analytical Approaches.}
An MDP with a continuous action space and a continuous state space can sometimes be solved via the partial derivatives of the Bellman Optimality Equation with respect to the action space \cite{bellman}.  This requires that the reward function  be expressable as a closed-form function of the current state $s_t$ and action $a_t$, which  in turn requires an analytical solution of the ODE system. An analytical solution may be possible when the ODE system is simple enough such as in the three-compartment SIR model of \cite{closed_sir}, but it is not possible in general. 
Dynamic programming approaches as well as Pontryagin’s maximum principle have also been used  in the optimal control of simple epidemiological models in \cite{Perkins, Lemecha}. 

\paragraph{Tree Search.}

Natively, Monte Carlo Tree Search (MCTS)~\cite{mcts0}
requires  both the state space and the action space to be discrete. 

Discretization of the action space requires domain knowledge, however. In our work with public health officials, we often found disagreement regarding the granularity and the values of such discretizations.

Moreover, MCTS  suffers from \textit{low sampling efficiency}. At each time step, MCTS estimates the value of a certain action through many Monte Carlo random plan simulations. The more simulations, the more accurate the estimate. This tradeoff implies that as the action space grows (with more interventions or with finer-grained discretizations), the computational resources have to  increase exponentially in order to obtain reasonable coverage. Without good coverage, plans generated by MCTS will be much more costly than simple hand-crafted plans.

\paragraph{Actor-Critic Methods.}  In contrast to the planning approaches discussed above, we consider an agent that produces a \textit{policy}.
    
\noindent \textbf{$\blacktriangleright$ Policy:} $\pi: \mathcal{S} \times \mathcal A \rightarrow [0, 1]$ is a probability distribution on the actions given a state $s$. 

In actor-critic methods, the agent is both a \textit{critic} that learns the  value of a specific state or an action, often with a neural network, and an \textit{actor}, that learns a policy, also through a neural network, as guided by the critic. The agent generates an intervention plan or schedule for an initial state $s_0$, by sequentially sampling actions $a_t \sim \pi(s_t)$, simulating the next state $s_{t+1} \sim p(s_t, a_t)$, and computing the reward $r_{t} = r(s_t,a_t,s_{t+1})$, repeatedly until the time horizon $T$.

Actor-Critic methods theoretically converge to the local maximum of the reward function \cite{actor-critic}. They inherently support continuous state and action spaces as the neural networks representing the actor and the critic support continuous input and output values. State-of-the-art actor-critic methods such as Soft Actor-Critic (SAC) \cite{sac} or Proximal Policy Optimization (PPO) \cite{ppo} have been shown to have good learning performance in a wide range of reinforcement learning tasks \cite{benchmark}. These methods, however, are sensitive to the setting of their hyperparameters, requiring extensive experimentation and benchmarking to appropriately tune them. In the following section, we test these methods on a new benchmark consisting of six different epidemic environments.

%% file: 04-eval.tex
\input{tables/actionspace-table}

\input{tables/baseline-table}

\input{tables/cost-table.tex}

\begin{figure*}[h]
\begin{subfigure}{0.33\textwidth}
  \centering
  \includegraphics[width=1\linewidth]{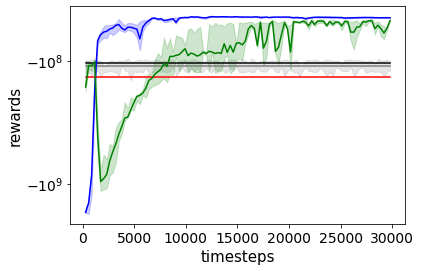}
  \caption{\sira}
  \label{fig:1}
\end{subfigure}%
\begin{subfigure}{0.33\textwidth}
  \centering
  \includegraphics[width=1\linewidth]{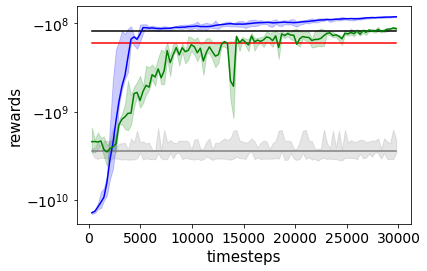}
  \caption{\sirva}
  \label{fig:3}
\end{subfigure}
\begin{subfigure}{0.33\textwidth}
  \centering
  \includegraphics[width=1\linewidth]{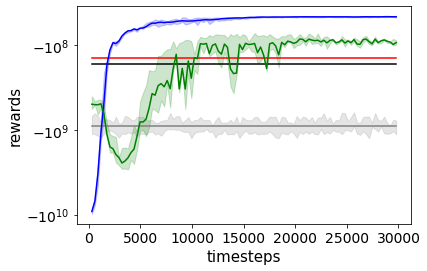}
  \caption{\cfa}
  \label{fig:5}
\end{subfigure}

\begin{subfigure}{0.33\textwidth}
  \centering
  \includegraphics[width=1\linewidth]{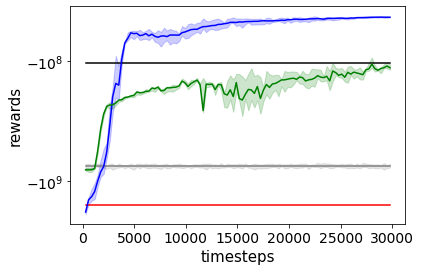}
  \caption{\sirb}
  \label{fig:2}
\end{subfigure}
\begin{subfigure}{0.33\textwidth}
  \centering
  \includegraphics[width=1\linewidth]{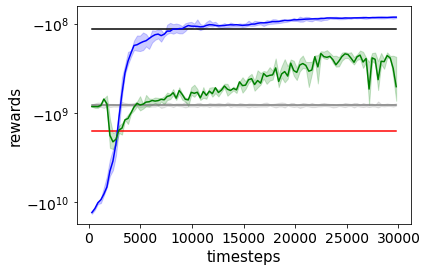}
  \caption{\sirvb}
  \label{fig:4}
\end{subfigure}
\begin{subfigure}{0.33\textwidth}
  \centering
  \includegraphics[width=1\linewidth]{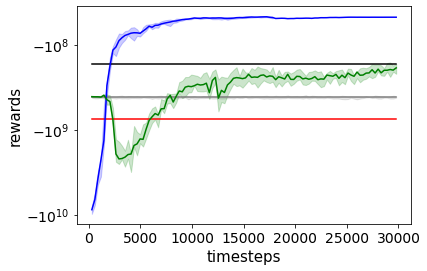}
  \caption{\cfb}
  \label{fig:6}
\end{subfigure}

\centering
\begin{subfigure}{0.5\textwidth}
  \centering
  \includegraphics[width=1\linewidth]{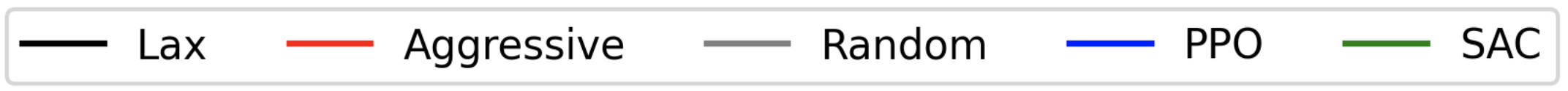}
  \label{fig:7}
\end{subfigure}

\caption{The cumulative reward (minimizing cost) of PPO and SAC over 30,000 training timesteps, averaged over four random seeds, We also plot the cumulative rewards of the three baseline policies, \textit{Lax}, \textit{Aggressive} and \textit{Random}. In all scenarios, PPO outperforms both SAC and the baseline policies.} 
\label{fig:learning}
\end{figure*}

\input{tables/results-table}

\section{Empirical Evaluation}
\label{sec:experiment}

\subsection{The Benchmark}
\label{sec:benchmark} 

The benchmark described here is largely inspired by a collaboration with a public health entity through a non-disclosure agreement that preserves their anonymity. Our public health colleagues wanted guidance on multiple interventions for mitigating different diseases. These interventions had control parameters that described the degree of their application.

\paragraph{State Spaces. }
We consider three compartmental models of increasing state-space complexity:
\begin{enumerate}
    \item \sir: The classic and basic three-compartment Susceptible (S), Infectious (I), and Recovered (R) model.
    \item \sirv: A modified SIR model with two vaccinated compartments for diseases having two-dose vaccination regimens. See Figure \ref{fig:sirv}.
    \item \cf: A 15-compartment model \cite{epipolicy}, which includes compartments for capturing different disease severity and symptoms,
     hospitalization 
    and isolation or quarantine.
    See Figure 5 in the Appendix.
\end{enumerate}

Both the \sirv and the \cf models capture reinfection by introducing a transition from the recovered (R) compartment to the susceptible (S) compartment. The \cf model also captures hospitalization and quarantine, which incurs an additional cost for states with hospitalized or quarantined individuals, while SIRV and SIR do not. The total population size is 2 million people with an initial infectious population of 0.005\% (100 infected individuals).

\paragraph{Action Spaces.}
For the action space, we consider four interventions, each of which has a continuous control parameter.  Table \ref{tab:interventions} describes these interventions and their control parameters. The table also describes two "action spaces" $A$ and $B$, each of which consists of certain interventions. 
A Reinforcement Learning agent learns a policy, which sets the values for each intervention's control parameters for a given state.

\paragraph{Environments.} We combine the three state spaces (\sir, \sirv, \cf) and the two  action spaces ($A$, $B$) to form six different benchmarking environments of varying disease and intervention complexity (\sira, \sirb, \sirva, \sirvb, \cfa, \cfb). The planning time-horizon is one year or 52 weeks with a one week timestep.

\paragraph{Baselines.} Table \ref{tab:baselines} describes plausible handmade baseline polices in terms of their intervention parameter settings. The \textit{Agressive} policy mimics plans where officials react early (first 120 days) and aggressively with school and workplace closures in the hopes of quickly curbing an epidemic and then relaxing these interventions. The \textit{Lax} policy mimics plans that favor less costly interventions such as mask-wearing mandates and a  relaxed schedule of vaccinations~\cite{policiesworld}. The \textit{Random} policy is one that chooses each control parameter value uniformly at random from its range.

\paragraph{RL Implementations.} We use Stable-Baselines3's implementations of SAC~\cite{JMLR} and a well-known PPO implementation~\cite{shengyi2022the37implementation}.  

\paragraph{Reward \& State Normalization.} 
As \citet{normalized} suggest, reward and state normalization are essential to achieve learning stability and convergence, especially when rewards or states can fluctuate by orders of magnitude across experiences. Thus, we normalize as follows: 

If $x$ is the current value for reward or for a state variable, then the normalized $\Tilde{x}$ is the z-score, viz. $\frac{x - \mu}{\sigma}$ where $\mu, \sigma$ are the running mean and standard deviation of $x$.

\paragraph{Hyperparameter Tuning.} For SAC, we use grid-search to find the best hyperparameter settings for the simplest environment (\sira). We then apply these settings to the other environments. Certain hyperparameters are set to their default values in the SAC implementation of Stable-Baseline3. For PPO, hyperparameters follow \citet{shengyi2022the37implementation}'s recommendations as the default Stable-Baseline3 PPO implementation failed to converge.

Table 6 in the Appendix lists all our hyperparameter values for PPO and SAC.

\paragraph{Training.} We train both algorithms for 30,000 timesteps on four random seeds: learning convergence occurred around 20,000 timesteps in all environments.

\paragraph{Hardware.} We conduct our experiments on Intel(R) Xeon(R) Gold 6230 CPU @ 2.10GHz with 80 cores.

\subsection{Results}
\label{sec:results}

Figure \ref{fig:learning} illustrates the learning performance of PPO and SAC over 30,000 timesteps on four different random seeds. Table \ref{tab:results} presents the highest cumulative rewards, averaged over the four seeds. We find that PPO significantly outperforms SAC and the baseline policies in all  six environments. SAC achieves similar rewards to PPO in the \sira and \sirva environments, but under-performs in the remaining four environments.

\subsubsection{Discussion \& Limitations} 
\begin{figure}[h]
  \centering
  \includegraphics[width=1\linewidth]{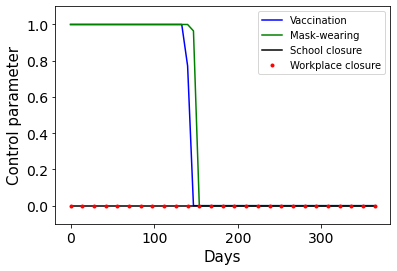}
  \caption{The highest-reward intervention schedule generated by PPO for the \sirvb environment. Mask-wearing and vaccination are aggressively used until the population is mostly vaccinated.} 
  \label{fig:plan}
\end{figure}

From these empirical results, we conclude that RL is a viable mechanism for generating alternative intervention plans that can reduce costs compared with plausible, hand-crafted policies. Figure  \ref{fig:plan} illustrates the schedule sampled from PPO's policy in the \sirvb environment. For PPO, which performs the best, we note that its generated plans are smooth with no wide fluctuations in control parameter settings, and can be easily described to policy-makers. For example the plan in figure \ref{fig:plan} is ``Enforce mask-wearing mandates and vaccinate as much as possible of the population, for almost 150 days. By then the population is mostly vaccinated or the pool of susceptible individuals is too small to cause a future outbreak. Then, relax these interventions. Do not enforce expensive interventions such as school or workplace closures for this particular disease environment. PPO suggests similar plans  for all the other environments (See Figure 6 in the Appendix.

While SAC performs as well as PPO in \sira and \sirva, it does not perform as well in the remaining environments. SAC is particularly sensitive to two hyperparameters: reward scale and the update frequency of the neural network. Since we tuned these SAC parameters only on the \sira environment, we suspect that each environment may require its own independent hyperparameter tuning. For the epidemic planning problem, this makes PPO a more robust alternative to SAC, one that requires less tuning or no tuning at all. We also note that SAC's high sample efficiency, when compared to PPO, may not be so important in our application due to the efficiency of the simulator. Finally, we note that training PPO for 30,000 timesteps takes significantly less wall-clock time when compared to SAC (e.g. 3 hours vs. 18 hours on our hardware per training run in \cfb). In critical situations, where policy-makers are exploring multiple different interventions to curb an ongoing epidemic, the time it takes for an RL agent to suggest intervention plans can make or break its integration into their policy-making workflows.

\subsubsection{Future Work.}

It is important to emphasize that the learned RL policies are heavily influenced by the  parameter values that define the disease model (e.g. fatality rate, recovery rate, etc) or the interventions (e.g. the cost of isolation, the effect of masks on transmissibility rate, etc). For example, reducing the cost of workplace closures, or increasing its effectiveness in reducing infectious interactions within workplaces in our benchmark may cause the RL agents to favor this intervention rather than excluding it. Inferring the exact value of these parameters, or calibration, is an active area of research~\cite{cali} and ~\cite{uncertain}.
Though we consider this to be independent of the optimization problem, it is possible to use our tool to examine the differences in plans sampled from the agent's policy when the disease has a high versus a low transmissibility rate, or when a vaccine has a high or low efficacy.

We have assumed the availability of a fully observable state in which we know the population size within every compartment. Future work can extend our work to partially observable MDPs where only partial state information is available. Finally, we note that we tuned our hyperparameters on a single configuration; in practice, hyperparameters should be tuned to be robust with respect to a range of parameter settings (e.g. different transition rates).

%% file: tables/actionspace-table.tex
\begin{table*}[h]
\centering
\resizebox{\textwidth}{!}{%
\begin{tabular}[t]{@{}ll|lccl|cc@{}}
\toprule
  \multicolumn{2}{c}{\textbf{Intervention}} &
  \multicolumn{4}{c}{\textbf{Control Parameter}} &
  \multicolumn{2}{c}{\textbf{Action Spaces}} \\[0.1cm] \midrule
 &
  \textbf{} &
  \textbf{} &
   \textbf{} &
  \textbf{Range} &
  \textbf{Description of control parameter's effect} &
  \textbf{A} &
  \textbf{B} \\[0.25cm]
1 &
  Mask-wearing &
  \begin{tabular}[t]{@{}l@{}}Compliance \\ Degree\end{tabular} &
  $m$ &
  [0, 1] &
  \begin{tabular}[t]{@{}l@{}}Affects the transition from S to I by \\ modifying the transmissibility $\beta$ rate. \\
  Higher compliance reduces the transmissibility rate \\
  as follows: $(1-Rm)\beta$, where $R \approx 0.8$ is the \\
  reduction in transmission rate at full compliance ($m=1$).\end{tabular} &
  Y &
  Y \\[1.75cm]
2 &
  Vaccination &
  \begin{tabular}[t]{@{}l@{}}Administration \\ Rate\end{tabular} &
  $v$ &
  [0, 1] &
  \begin{tabular}[t]{@{}l@{}}Moves $vC$ individuals from S to R in \sir, \\ S to V1 in \sirv and S to V in \cf. \\ $C \approx 10,000$ is the maximum number of doses available. 
  \end{tabular} &
  Y &
  Y \\[1cm]
3 &
  School closure &
  \begin{tabular}[t]{@{}l@{}}Remote learning \\ proportion\end{tabular} &
  $s$ &
  [0, 1] &
  \begin{tabular}[t]{@{}l@{}}Higher proportions lower infectious interactions within \\ school facilities; \\$s=0$ means no remote learning/school closure.\end{tabular} &
   &
  Y\\[1cm]
4 &
  Workplace closure &
  \begin{tabular}[t]{@{}l@{}}Remote working \\ proportion\end{tabular} &
  $w$ &
  [0, 1] &
  \begin{tabular}[t]{@{}l@{}}Higher proportions lower infectious interactions within \\ workplace facilities; \\ $w=0$ means no workplace closures.\end{tabular} &
   &
  Y \\
  \bottomrule
\end{tabular}%
}
\caption{The four interventions in the different action spaces $A, B$. The costs associated with each intervention can be found in table \ref{tab:cost}.}
\label{tab:interventions}
\end{table*}

%% file: tables/baseline-table.tex
\begin{table}[h]
\centering
\resizebox{\linewidth}{!}{%
\begin{tabular}{@{}cccc@{}}
\toprule
  \textbf{} &
  \multicolumn{3}{c}{\textbf{Baseline Policies}} \\
\midrule
  \textbf{} &
  \multicolumn{1}{c}{\textbf{Aggressive}} &
  \multicolumn{1}{c}{\textbf{Lax}} &
  \multicolumn{1}{c}{\textbf{Random}} \\

  $m$ &
  \multicolumn{1}{c}{0.8} &
  \multicolumn{1}{c}{1} &
  $\sim U(0,1)$ \\

  $v$ &
  \begin{tabular}[c]{@{}c@{}}85\% population in\\ 9 months\end{tabular} &
  \begin{tabular}[c]{@{}c@{}}70\% population in\\ 12 months\end{tabular} &
  $\sim U(0,1)$ \\

  $s$ &
  \begin{tabular}[c]{@{}c@{}}if $t \in [0, 120]$ then 1, \\ 0.5 otherwise;\end{tabular} &
  0 &
  $\sim U(0,1)$ \\

  $w$ &
  \begin{tabular}[c]{@{}c@{}}if $t \in [0, 120]$ then 1, \\ 0.5 otherwise;\end{tabular} &
  0 &
  $\sim U(0,1)$\\
  \bottomrule
\end{tabular}%
}
\caption{The \textit{Aggressive} and the \textit{Lax} baseline policies have set control parameter settings for the interventions, while the values for the \textit{Random} baseline are sampled uniformly at random from the parameter's range.}

\label{tab:baselines}
\end{table}

%% file: tables/cost-table.tex
\begin{table}[h]
\begin{tabular}{ll}
\toprule
\textbf{Intervention}  & \textbf{Cost}      \\ \midrule
 Mask-wearing                                & 0.05\$ per person wearing mask per day \\
 Vaccination                                 & 40\$ per person getting vaccinated     \\
 School closure                              & 1.8\$ per affected person per day      \\
 Workplace closure                           & 1.8\$ per affected person per day      \\ \midrule

 \textbf{Disease State} & \textbf{Cost}     \\ \midrule
 Infections                                  & 173\$ per infectious person per day    \\
 Hospitalizations                            & 250\$ per hospitalized person per day  \\
 Fatalities                                  & 100,000\$ for each fatality            \\ \bottomrule
\end{tabular}
\caption{The four interventions under consideration with their respective cost as well as the disease burden cost.}
\label{tab:cost}
\end{table}

%% file: tables/results-table.tex
\begin{table*}[htb]
\resizebox{\textwidth}{!}{%
\begin{tabular}{@{}ll|cc|cc|cc@{}}
\toprule
\multicolumn{2}{c|}{Policy} &
  \multicolumn{1}{c}{\sira} &
  \multicolumn{1}{c|}{\sirb} &
  \multicolumn{1}{c}{\sirva} &
  \multicolumn{1}{c|}{\sirvb} &
  \multicolumn{1}{c}{\cfa} &
  \multicolumn{1}{c}{\cfb} \\ \midrule

Random & $\max$ &$-94.1$ &
$-693.5$
& $-1008.3$
& $-717.4$
& $-559.3$
& $-387.7$\\
 & $\mu\pm\sigma$   & $-110.4 \pm 20$ & $-745.5 \pm 46.6$ & $-2798.3 \pm 1049$  & $-811.9 \pm 61.8$ & $-905.2 \pm 291$ & $-415.4 \pm 20.5$ \\[0.15cm]

{Aggressive} & & $-135.9$        & $-1576$           & $-166.3$            & $-1577.1$         & $-139.9$         & $-742.9$          \\[0.15cm]

{Lax} & & $-104.4$        & $-104.2$          & $-122.4$            & $-115.6$          & $-167.6$         & $-166.6$          \\[0.15cm]

SAC & $\max$&
$-43.4$&
$-73.4$&
$-99.9$&
$-169.2$&
$-65.6$&
$-128.6$\\
& $\mu\pm\sigma$     &  $-45.3 \pm 1.9$	& $-92.2 \pm 12.6$	& $-108.6 \pm 5.7$	& $-200.4 \pm 22.7$	& $-73.7 \pm 7.2$	& $-176.7 \pm 53.7$ \\[0.15cm]

PPO& $\max$&
$-42.9$&
$-42.1$&
$-82.2$&
$-78.9$&
$-45.3$&
$-45.2$\\
&$\mu\pm\sigma$       & $-43.3 \pm 0.2$ & $-42.5 \pm 0.3$   & $-83.4 \pm 1.2$     & $-81.2 \pm 1.5$   & $-45.7 \pm 0.47$ & $-46.2 \pm 0.52$  \\ \bottomrule
\end{tabular}%
}
\caption{Maximum, mean and standard deviation of the highest cumulative rewards (expressed as negative costs) achieved by each reinforcement learning algorithm or baseline policy across four different random seeds. All maximum and mean rewards and standard deviations are in millions  of dollars ($10^6$).}
\label{tab:results}
\end{table*}

%% file: 07-related-works.tex
\section{Related Work}

\paragraph{Historical Approaches.}

Using Markov Decision Processes (MDPs) for epidemic planning was explored as early as 1981 by \citet{lev}. \citet{lev} analytically proved that optimal levels of quarantine or medical care are non-decreasing functions of the size of the infectious population. More recently, research on epidemic control has shifted towards a more computational approach where simulations are carefully designed and analyzed to provide insights for epidemic planning. For example, \citet{savi, klec, ole}, compare, through simulations, many well-established intervention plans. 
 
Our work complements this body of research by demonstrating reinforcement learning as a viable approach to generate alternative candidate strategies that are tailored to a specific environment. 

\paragraph{Reinforcement Learning for Epidemic Control.} The COVID pandemic   inspired new research into using RL approaches for epidemic control. 
Among recent works that have formulated different MDPs with actions capturing interventions with the goal of using RL to inform epidemic control, we found \cite{EpOp, Varun, Pieter} to be the most closely related to our work.

While our motivations are similar, we differ in that  we have demonstrated RL as a viable approach to generate intervention plans when there are \textit{multiple, continuously-parameterizable interventions} that can influence different aspects of the underlying disease and population model. By contrast, \citet{EpOp, Varun} combine the effect of many interventions into one that reduces the transmission rate of the disease, thus controlling the disease only via this single parameter. \citet{Pieter} uses RL to study the cost-effectiveness of different degrees of school closures across regions. In our case, the programmatically-defined interventions can modify any transition rate, and even redistribute the population across disease compartments, regions or facilities. Moreover, as the RL agent can control each intervention independently, it can determine the tradeoffs of different interventions in terms of their costs and benefits ---  an option that is not possible when grouping multiple interventions into one.

Table \ref{tab:related_works} provides a feature comparison of the RL approaches used in many of the recent works for epidemic planning.

\input{tables/related-works-table.tex}

%% file: tables/related-works-table.tex
\begin{table}[h]
\resizebox{\linewidth}{!}{%
\begin{tabular}[t]{lccccc}


\toprule
 
& \multicolumn{2}{c}{$\mathcal{S}$}
& \multicolumn{2}{c}{$\mathcal A$}
&  \\
\textbf{Related Works} & C & L & C & L & {Sim.}\\
\midrule 

\begin{tabular}[b]{@{}l@{}}\citet{bastani2021efficient,abdallah2022deep} 
\end{tabular}
& \demono & \demono & \demono & \demono & \demono \\[0.5em] 

\begin{tabular}[b]{@{}l@{}}
\cite{Pieter, khadilkar2020optimising} \\
\citet{zong2022reinforcement, du2022district}\\
\citet{arango2020covid}
\end{tabular} 
& \demoyes & \demoyes & \demono & \demono & \demono \\[0.5em]  

\begin{tabular}[b]{@{}l@{}}
\citet{Varun,EpOp}\\
\citet{feng2022precise, bampa2022learning}\\
\citet{ohi2020exploring,jiang2020epidemic} 
\end{tabular} 
& \demoyes & \demoyes & \demono & \demono & \demoyes \\[0.5em] 

\citet{song2020reinforced} & \demoyes & \demoyes & \demoyes & \demono & \demono \\

\citet{liu2020microscopic} & \demono & \demono & \demoyes & \demoyes & \demono \\

\citet{chadi2022reinforcement} & \demoyes & \demoyes & \demono & \demoyes & \demono \\[0.5em]

\begin{tabular}[b]{@{}l@{}}
\citet{kwak2021deep,feng2022contact} \\
\citet{bushaj2022simulation}
\end{tabular} 
& \demoyes & \demoyes & \demono & \demoyes & \demoyes \\[0.5em]

\citet{padmanabhan2021reinforcement} & \demoyes & \demoyes & \demoyes & \demoyes & \demono \\ 

\midrule

\textbf{Our Work} & \textbf{\demoyes} & \textbf{\demoyes} & \textbf{\demoyes} & \textbf{\demoyes} & \textbf{\demoyes} \\ 

\bottomrule 
\end{tabular}
}
\caption{A comparison of recent RL epidemic planning tools in terms of the size of the supported state space $\mathcal S$ and action space $\mathcal A$ (large (L) \demoyes~ or small/singular \demono) and their continuity (continuous (C) \demoyes~ or discrete \demono), and the integration of an epidemic simulator (Sim.) that can simulate different diseases. 
} 
\label{tab:related_works}
\end{table}

%% file: 06-conclusion.tex
\section{Conclusions} 

We present an approach to support public health officials in their efforts to combat epidemics. The approach consists of a reinforcement learning algorithm that proposes policy choices consisting of multiple continuous interventions (e.g., masking, vaccination, isolation, and others, all to various degrees). The ability to model multiple interventions is important, because policy-makers often have multiple strategies available. 
The ability to model each intervention continuously is important because it is often impractical to impose an intervention absolutely (e.g. critical workers may need to interact with one another). 
Our tool offers policy-makers quantitatively backed recommendations  while leaving to them the final choice of a strategy, which may involve intangibles such as culture or politics. 

Our benchmarking environments are similar to real-world epidemic planning environments both in terms of disease models used (\sir, \sirv and \cf) and interventions considered. In addition to its application to epidemiology,  our work advances reinforcement learning research, through the provision of a realistic benchmarking environment.

\section*{Acknowledgments} This work was supported by the NYUAD Center for Interacting Urban Networks (CITIES), funded by Tamkeen under the NYUAD Research Institute Award CG001, by the NYUAD COVID-19 Facilitator Research Fund, and by the ASPIRE Award for Research Excellence (AARE-2020) grant AARE20-307.

%% file: 08-appendix.tex
\section{Appendix}
\begin{figure*}[htb]
    \centering
    \includegraphics[width=1\textwidth]{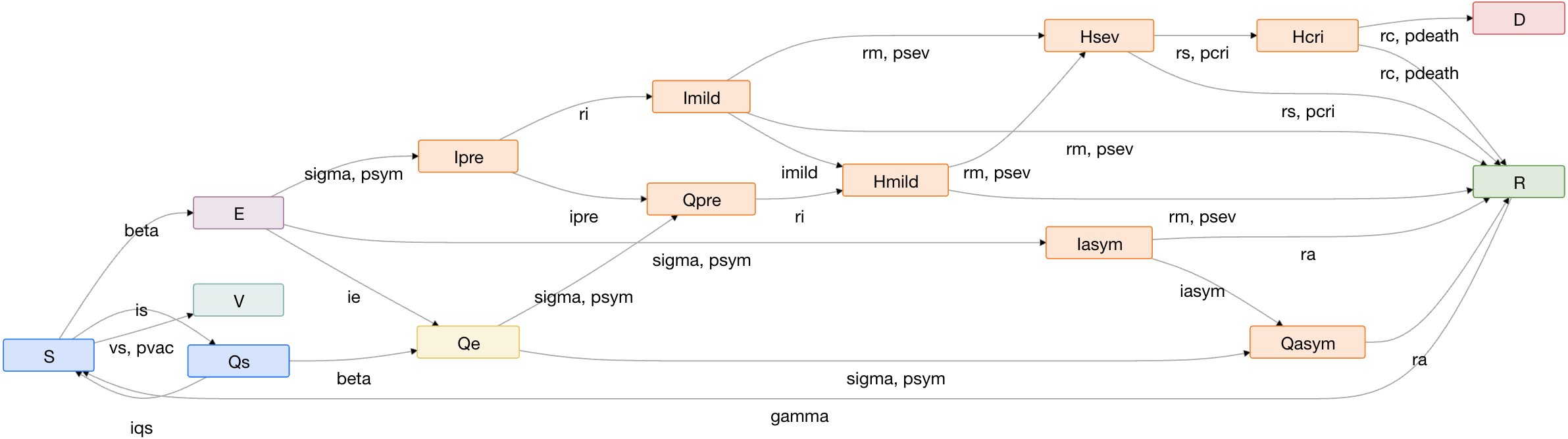}
    \caption{\cf: A compartmental model with 15 compartments that affords several interventions such as vaccinations, school and workplace closures.}
    \label{fig:covid19}
\end{figure*}

\begin{figure*}[htb]
\begin{subfigure}{0.33\textwidth}
  \centering
  \includegraphics[width=1\linewidth]{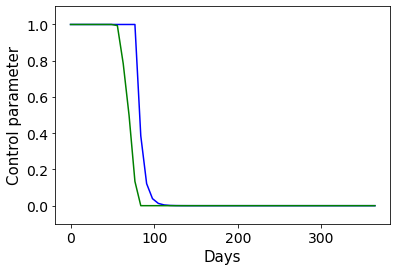}
  \caption{\sira}
  \label{fig:i1}
\end{subfigure}%
\begin{subfigure}{0.33\textwidth}
  \centering
  \includegraphics[width=1\linewidth]{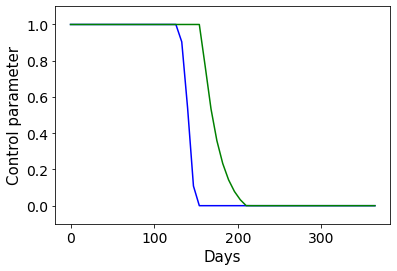}
  \caption{\sirva}
  \label{fig:i2}
\end{subfigure}
\begin{subfigure}{0.33\textwidth}
  \centering
  \includegraphics[width=1\linewidth]{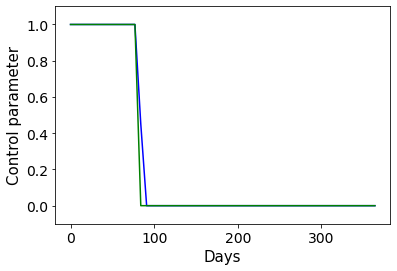}
  \caption{\cfa}
  \label{fig:i3}
\end{subfigure}

\begin{subfigure}{0.33\textwidth}
  \centering
  \includegraphics[width=1\linewidth]{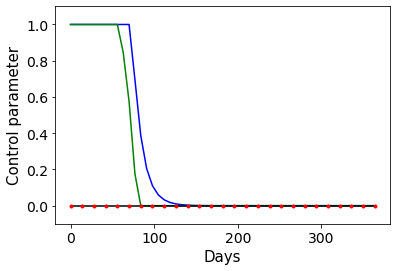}
  \caption{\sirb}
  \label{fig:i4}
\end{subfigure}%
\begin{subfigure}{0.33\textwidth}
  \centering
  \includegraphics[width=1\linewidth]{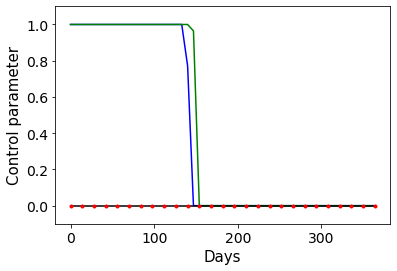}
  \caption{\sirvb}
  \label{fig:i5}
\end{subfigure}
\begin{subfigure}{0.33\textwidth}
  \centering
  \includegraphics[width=1\linewidth]{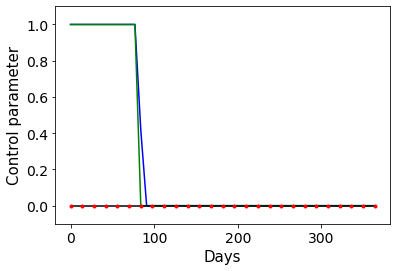}
  \caption{\cfb}
  \label{fig:i6}
\end{subfigure}

\centering
\begin{subfigure}{0.5\textwidth}
  \centering
  \includegraphics[width=1\linewidth]{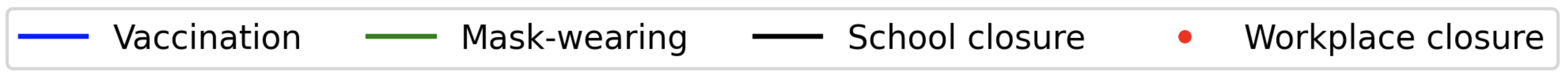}
  \label{fig:7}
\end{subfigure}
\caption{Intervention schedules generated by PPO in all 6 environments.}
\label{fig:plans_PPO}
\end{figure*}

\begin{figure*}[htb]
\begin{subfigure}{0.33\textwidth}
  \centering
  \includegraphics[width=1\linewidth]{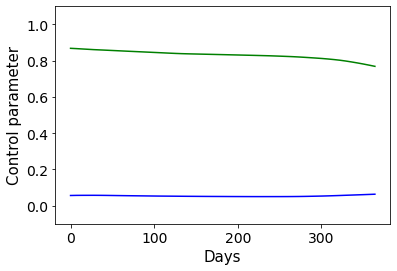}
  \caption{\sira}
  \label{fig:i1}
\end{subfigure}%
\begin{subfigure}{0.33\textwidth}
  \centering
  \includegraphics[width=1\linewidth]{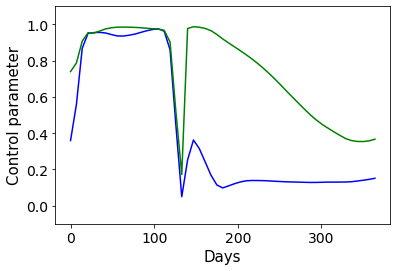}
  \caption{\sirva}
  \label{fig:i2}
\end{subfigure}
\begin{subfigure}{0.33\textwidth}
  \centering
  \includegraphics[width=1\linewidth]{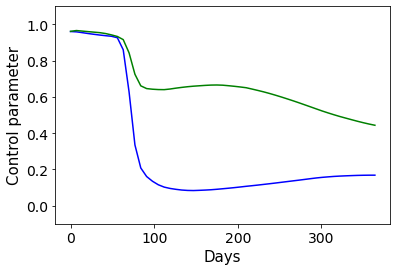}
  \caption{\cfa}
  \label{fig:i3}
\end{subfigure}

\begin{subfigure}{0.33\textwidth}
  \centering
  \includegraphics[width=1\linewidth]{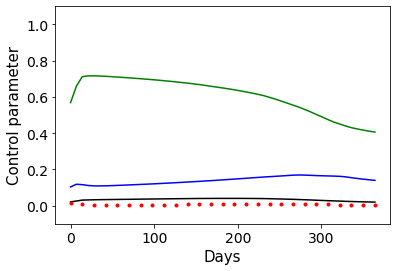}
  \caption{\sirb}
  \label{fig:i4}
\end{subfigure}%
\begin{subfigure}{0.33\textwidth}
  \centering
  \includegraphics[width=1\linewidth]{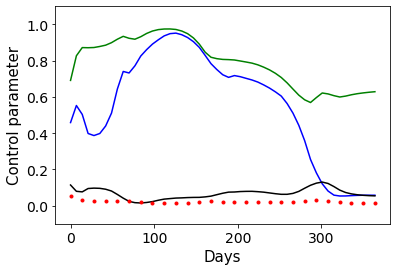}
  \caption{\sirvb}
  \label{fig:i5}
\end{subfigure}
\begin{subfigure}{0.33\textwidth}
  \centering
  \includegraphics[width=1\linewidth]{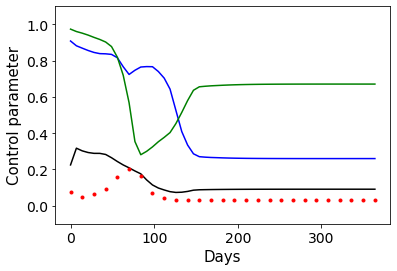}
  \caption{\cfb}
  \label{fig:i6}
\end{subfigure}

\centering
\begin{subfigure}{0.5\textwidth}
  \centering
  \includegraphics[width=1\linewidth]{figures/intervention_schedule_legend.png}
  \label{fig:8}
\end{subfigure}
\caption{Intervention schedules generated by SAC in all 6 environments }
\label{fig:plans_SAC}
\end{figure*}

This appendix includes the 15-compartment model used in the \cf benchmarking environments (Figure \ref{fig:covid19}), the RL hyperparameter settings  (Table \ref{tab:hyperparameter}), and the highest cumulative reward intervention plans generated by PPO and SAC (Figures \ref{fig:plans_PPO} and \ref{fig:plans_SAC}). 

\begin{table*}[htb]
\centering
\resizebox{\textwidth}{!}{
\begin{tabular}[t]{llcc}
\toprule
\textbf{SAC Hyperparameter} & 
\multicolumn{1}{c}{\textbf{Description}}
& \textbf{Grid-search values}  
& \textbf{Chosen value} \\ 
\midrule

Reward scale                                    
 & \begin{tabular}[t]{@{}l@{}}This is the multiplicative scaling applied after the normalized rewards. \\ It is equivalent to inverse of entropy regularization coefficient\end{tabular} 
 & \begin{tabular}[t]{@{}c@{}}{[}1, 10, 100, \\1000{]}  \end{tabular}      
 & 100                   \\[0.15cm]

Learning rate                                   
& \begin{tabular}[t]{@{}l@{}}Learning rate for Adam optimizer, the same learning rate \\ will be used for all networks (Q-Values, Actor and Value function)\end{tabular}                
& \begin{tabular}[t]{@{}c@{}}{[}0.003, 0.0003,\\ 0.00003{]} \end{tabular}  
& 0.0003                \\[0.15cm]

Training frequency                             
& The number of simulation timesteps to update the model
& {[}5, 25, 125{]}             
& 5                     \\[0.15cm]

Learning starts                                 
& \begin{tabular}[t]{@{}l@{}}How many steps of the model to collect transitions for before \\ learning starts\end{tabular} 
& \begin{tabular}[t]{@{}c@{}}{[}1000, 5000, \\10000{]} \end{tabular}     
& 1000                  \\[0.15cm]

Batch size
& Minibatch size for each gradient update   
&                              
& 256                   \\[0.15cm]

Tau                                             
& The soft update coefficient (``Polyak update", between 0 and 1)            &                             
& 0.005                 \\[0.15cm]

Gamma                                           
& The discount factor                                                        &                             
& 0.99                  \\[0.15cm]

Buffer size                                     
& Size of the replay buffer 
&
& 1000000               \\[0.15cm]

Gradient steps                                  
& How many gradient steps to do after each rollout     
&                              
& 1                     
\\ \midrule

\textbf{PPO Hyperparameter}
& 
&
&
\\ \midrule

Learning rate                                   
& The learning rate of the optimizer
& 
&0.0003\\[0.15cm]

Num steps                                       
& The number of steps to run in each environment per policy rollout   
& 
&2048\\[0.15cm]

Gamma                                           
& The discount factor
& 
&0.99\\[0.15cm]

Gae Lambda
& The lambda for the general advantage estimation                            & 
&0.95\\[0.15cm]

Num minibatches 
& The number of mini-batches                                                 & 
&32\\[0.15cm]

Update epochs 
& The K epochs to update the policy                                          &
&10\\[0.15cm]

Clip coefficient
& The surrogate clipping coefficient
&
&0.2\\[0.15cm]

Entropy coefficient
& Coefficient of the entropy
&
&0.0\\[0.15cm]

Vf coefficient
& Coefficient of the value function
&
&0.5\\[0.15cm]

Max grad norm
& The maximum norm for the gradient clipping                                 
&
&0.5\\
\bottomrule
\end{tabular}
}
\caption{The hyperparameter settings for SAC and PPO.}
\label{tab:hyperparameter}
\end{table*}

%% file: ijcai23.bbl
\begin{thebibliography}{}

\bibitem[\protect\citeauthoryear{Abdallah \bgroup \em et al.\egroup
  }{2022}]{abdallah2022deep}
Wejden Abdallah, Dalel Kanzari, Dorsaf Sallami, Kurosh Madani, and Khaled
  Ghedira.
\newblock A deep reinforcement learning based decision-making approach for
  avoiding crowd situation within the case of covid'19 pandemic.
\newblock {\em Computational Intelligence}, 38(2):416--437, 2022.

\bibitem[\protect\citeauthoryear{Andrychowicz \bgroup \em et al.\egroup
  }{2020}]{normalized}
Marcin Andrychowicz, Anton Raichuk, Piotr Stanczyk, Manu Orsini, Sertan Girgin,
  Rapha{\"{e}}l Marinier, L{\'{e}}onard Hussenot, Matthieu Geist, Olivier
  Pietquin, Marcin Michalski, Sylvain Gelly, and Olivier Bachem.
\newblock What matters in on-policy reinforcement learning? {A} large-scale
  empirical study.
\newblock {\em CoRR}, abs/2006.05990, 2020.

\bibitem[\protect\citeauthoryear{Arango and Pelov}{2020}]{arango2020covid}
Mauricio Arango and Lyudmil Pelov.
\newblock Covid-19 pandemic cyclic lockdown optimization using reinforcement
  learning.
\newblock {\em arXiv preprint arXiv:2009.04647}, 2020.

\bibitem[\protect\citeauthoryear{Bampa \bgroup \em et al.\egroup
  }{2022}]{bampa2022learning}
Maria Bampa, Tobias Fasth, Sindri Magnusson, and Panagiotis Papapetrou.
\newblock : Learning intervention strategies for epidemics with reinforcement
  learning.
\newblock In {\em International Conference on Artificial Intelligence in
  Medicine}, pages 189--199. Springer, 2022.

\bibitem[\protect\citeauthoryear{Barlow and Weinstein}{2020}]{closed_sir}
N.~S. Barlow and S.~J. Weinstein.
\newblock {{A}ccurate closed-form solution of the {S}{I}{R} epidemic model}.
\newblock {\em Physica D}, 408:132540, Jul 2020.

\bibitem[\protect\citeauthoryear{Bastani \bgroup \em et al.\egroup
  }{2021}]{bastani2021efficient}
Hamsa Bastani, Kimon Drakopoulos, Vishal Gupta, Ioannis Vlachogiannis, Christos
  Hadjichristodoulou, Pagona Lagiou, Gkikas Magiorkinis, Dimitrios Paraskevis,
  and Sotirios Tsiodras.
\newblock Efficient and targeted covid-19 border testing via reinforcement
  learning.
\newblock {\em Nature}, 599(7883):108--113, 2021.

\bibitem[\protect\citeauthoryear{Bushaj \bgroup \em et al.\egroup
  }{2022}]{bushaj2022simulation}
Sabah Bushaj, Xuecheng Yin, Arjeta Beqiri, Donald Andrews, and {\.I}~Esra
  B{\"u}y{\"u}ktahtak{\i}n.
\newblock A simulation-deep reinforcement learning (sirl) approach for epidemic
  control optimization.
\newblock {\em Annals of Operations Research}, pages 1--33, 2022.

\bibitem[\protect\citeauthoryear{Chadi and
  Mousannif}{2022}]{chadi2022reinforcement}
Mohamed-Amine Chadi and Hajar Mousannif.
\newblock A reinforcement learning based decision support tool for epidemic
  control: Validation study for covid-19.
\newblock {\em Applied Artificial Intelligence}, pages 1--33, 2022.

\bibitem[\protect\citeauthoryear{Colas \bgroup \em et al.\egroup }{2020}]{EpOp}
C{\'{e}}dric Colas, Boris~P. Hejblum, S{\'{e}}bastien Rouillon, Rodolphe
  Thi{\'{e}}baut, Pierre{-}Yves Oudeyer, Cl{\'{e}}ment Moulin{-}Frier, and
  M{\'{e}}lanie Prague.
\newblock Epidemioptim: {A} toolbox for the optimization of control policies in
  epidemiological models.
\newblock {\em CoRR}, abs/2010.04452, 2020.

\bibitem[\protect\citeauthoryear{Du \bgroup \em et al.\egroup
  }{2022}]{du2022district}
Xinqi Du, Tianyi Liu, Songwei Zhao, Jiuman Song, and Hechang Chen.
\newblock District-coupled epidemic control via deep reinforcement learning.
\newblock In {\em International Conference on Knowledge Science, Engineering
  and Management}, pages 417--428. Springer, 2022.

\bibitem[\protect\citeauthoryear{Feng \bgroup \em et al.\egroup
  }{2022a}]{feng2022contact}
Tao Feng, Sirui Song, Tong Xia, and Yong Li.
\newblock Contact tracing and epidemic intervention via deep reinforcement
  learning.
\newblock {\em ACM Transactions on Knowledge Discovery from Data (TKDD)}, 2022.

\bibitem[\protect\citeauthoryear{Feng \bgroup \em et al.\egroup
  }{2022b}]{feng2022precise}
Tao Feng, Tong Xia, Xiaochen Fan, Huandong Wang, Zefang Zong, and Yong Li.
\newblock Precise mobility intervention for epidemic control using unobservable
  information via deep reinforcement learning.
\newblock In {\em Proceedings of the 28th ACM SIGKDD Conference on Knowledge
  Discovery and Data Mining}, pages 2882--2892, 2022.

\bibitem[\protect\citeauthoryear{Ferguson \bgroup \em et al.\egroup
  }{2020}]{feg}
Neil Ferguson, Daniel Laydon, Gemma Nedjati-Gilani, and et~al.
\newblock Impact of non-pharmaceutical interventions (npis) to reduce covid-19
  mortality and healthcare demand.
\newblock 2020.

\bibitem[\protect\citeauthoryear{Haarnoja \bgroup \em et al.\egroup
  }{2018}]{sac}
Tuomas Haarnoja, Aurick Zhou, Pieter Abbeel, and Sergey Levine.
\newblock Soft actor-critic: Off-policy maximum entropy deep reinforcement
  learning with a stochastic actor.
\newblock {\em CoRR}, abs/1801.01290, 2018.

\bibitem[\protect\citeauthoryear{Hale \bgroup \em et al.\egroup
  }{2021}]{policiesworld}
Thomas Hale, Noam Angrist, Rafael Goldszmidt, Beatriz Kira, Anna Petherick,
  Toby Phillips, Samuel Webster, Emily Cameron-Blake, Laura Hallas, Saptarshi
  Majumdar, and Helen Tatlow.
\newblock A global panel database of pandemic policies (oxford covid-19
  government response tracker).
\newblock {\em Nature Human Behaviour}, 5, 04 2021.

\bibitem[\protect\citeauthoryear{Hazelbag \bgroup \em et al.\egroup
  }{2020}]{cali}
C~Marijn Hazelbag, Jonathan Dushoff, Emanuel~M Dominic, Zinhle~E Mthombothi,
  and Wim Delva.
\newblock Calibration of individual-based models to epidemiological data: A
  systematic review.
\newblock {\em PLoS Comput. Biol.}, 16(5):e1007893, May 2020.

\bibitem[\protect\citeauthoryear{Huang \bgroup \em et al.\egroup
  }{2022}]{shengyi2022the37implementation}
Shengyi Huang, Rousslan Fernand~Julien Dossa, Antonin Raffin, Anssi Kanervisto,
  and Weixun Wang.
\newblock The 37 implementation details of proximal policy optimization.
\newblock In {\em ICLR Blog Track}, 2022.
\newblock
  https://iclr-blog-track.github.io/2022/03/25/ppo-implementation-details/.

\bibitem[\protect\citeauthoryear{Jiang \bgroup \em et al.\egroup
  }{2020}]{jiang2020epidemic}
Yuanshuang Jiang, Linfang Hou, Yuxiang Liu, Zhuoye Ding, Yong Zhang, and
  Shengzhong Feng.
\newblock Epidemic control based on reinforcement learning approaches.
\newblock 2020.

\bibitem[\protect\citeauthoryear{Karim~et al.}{2021}]{covid-variant}
S.S.A Karim~et al.
\newblock {{O}micron {S}{A}{R}{S}-{C}o{V}-2 variant: a new chapter in the
  {C}{O}{V}{I}{D}-19 pandemic}.
\newblock {\em Lancet}, 398(10317):2126--2128, 12 2021.

\bibitem[\protect\citeauthoryear{Khadilkar \bgroup \em et al.\egroup
  }{2020}]{khadilkar2020optimising}
Harshad Khadilkar, Tanuja Ganu, and Deva~P Seetharam.
\newblock Optimising lockdown policies for epidemic control using reinforcement
  learning.
\newblock {\em Transactions of the Indian National Academy of Engineering},
  5(2):129--132, 2020.

\bibitem[\protect\citeauthoryear{Kleczkowski \bgroup \em et al.\egroup
  }{2012}]{klec}
Adam Kleczkowski, Katarzyna Ole{\'s}, Ewa Gudowska-Nowak, and Christopher~A
  Gilligan.
\newblock Searching for the most cost-effective strategy for controlling
  epidemics spreading on regular and small-world networks.
\newblock {\em J. R. Soc. Interface}, 9(66):158--169, January 2012.

\bibitem[\protect\citeauthoryear{Kocsis and Szepesvári}{2006}]{mcts0}
Levente Kocsis and Csaba Szepesvári.
\newblock Bandit based monte-carlo planning.
\newblock volume 2006, pages 282--293, 09 2006.

\bibitem[\protect\citeauthoryear{Kompella \bgroup \em et al.\egroup
  }{2020}]{Varun}
Varun Kompella, Roberto Capobianco, Stacy Jong, Jonathan Browne, Spencer~J.
  Fox, Lauren~Ancel Meyers, Peter~R. Wurman, and Peter Stone.
\newblock Reinforcement learning for optimization of {COVID-19} mitigation
  policies.
\newblock {\em CoRR}, abs/2010.10560, 2020.

\bibitem[\protect\citeauthoryear{Konda and Tsitsiklis}{1999}]{actor-critic}
Vijay Konda and John Tsitsiklis.
\newblock Actor-critic algorithms.
\newblock In S.~Solla, T.~Leen, and K.~M\"{u}ller, editors, {\em Advances in
  Neural Information Processing Systems}, volume~12. MIT Press, 1999.

\bibitem[\protect\citeauthoryear{Kwak \bgroup \em et al.\egroup
  }{2021}]{kwak2021deep}
Gloria~Hyunjung Kwak, Lowell Ling, and Pan Hui.
\newblock Deep reinforcement learning approaches for global public health
  strategies for covid-19 pandemic.
\newblock {\em PloS one}, 16(5):e0251550, 2021.

\bibitem[\protect\citeauthoryear{Lee \bgroup \em et al.\egroup }{2019}]{eco}
Ye-Rin Lee, Bogeum Cho, Min-Woo Jo, Minsu Ock, Donghoon Lee, Doungkyu Lee,
  Moon~Jung Kim, and In-Hwan Oh.
\newblock Measuring the economic burden of disease and injury in korea, 2015.
\newblock {\em Journal of Korean medical science}, 34(Suppl 1):e80--e80, Mar
  2019.
\newblock 30923489[pmid].

\bibitem[\protect\citeauthoryear{Lefèvre}{1981}]{lev}
Claude Lefèvre.
\newblock Optimal control of a birth and death epidemic process.
\newblock {\em Operations Research}, 29(5):971--982, 1981.

\bibitem[\protect\citeauthoryear{Libin \bgroup \em et al.\egroup
  }{2020}]{Pieter}
Pieter Libin, Arno Moonens, Timothy Verstraeten, Fabian Perez{-}Sanjines, Niel
  Hens, Philippe Lemey, and Ann Now{\'{e}}.
\newblock Deep reinforcement learning for large-scale epidemic control.
\newblock {\em CoRR}, abs/2003.13676, 2020.

\bibitem[\protect\citeauthoryear{Littman \bgroup \em et al.\egroup }{2013}]{dp}
Michael~L. Littman, Thomas~L. Dean, and Leslie~Pack Kaelbling.
\newblock On the complexity of solving markov decision problems.
\newblock {\em CoRR}, abs/1302.4971, 2013.

\bibitem[\protect\citeauthoryear{Liu}{2020}]{liu2020microscopic}
Changliu Liu.
\newblock A microscopic epidemic model and pandemic prediction using
  multi-agent reinforcement learning.
\newblock {\em arXiv preprint arXiv:2004.12959}, 2020.

\bibitem[\protect\citeauthoryear{Maharaj and Kleczkowski}{2012}]{savi}
Savi Maharaj and Adam Kleczkowski.
\newblock Controlling epidemic spread by social distancing: do it well or not
  at all.
\newblock {\em BMC Public Health}, 12(1):679, August 2012.

\bibitem[\protect\citeauthoryear{Mai \bgroup \em et al.\egroup }{2022}]{xrds}
Anh Le~Xuan Mai, Miro Mannino, Zain Tariq, Azza Abouzied, and Dennis Shasha.
\newblock Epipolicy: A tool for combating epidemics.
\newblock {\em XRDS}, 28(2):24–29, jan 2022.

\bibitem[\protect\citeauthoryear{Obsu and Balcha}{2020}]{Lemecha}
Legesse~Lemecha Obsu and Shiferaw~Feyissa Balcha.
\newblock Optimal control strategies for the transmission risk of covid-19.
\newblock {\em Journal of Biological Dynamics}, 14(1):590--607, 2020.
\newblock PMID: 32696723.

\bibitem[\protect\citeauthoryear{Ohi \bgroup \em et al.\egroup
  }{2020}]{ohi2020exploring}
Abu~Quwsar Ohi, MF~Mridha, Muhammad~Mostafa Monowar, Md~Hamid, et~al.
\newblock Exploring optimal control of epidemic spread using reinforcement
  learning.
\newblock {\em Scientific reports}, 10(1):1--19, 2020.

\bibitem[\protect\citeauthoryear{Ole{\'s} \bgroup \em et al.\egroup
  }{2012}]{ole}
Katarzyna Ole{\'s}, Ewa Gudowska-Nowak, and Adam Kleczkowski.
\newblock Understanding disease control: influence of epidemiological and
  economic factors.
\newblock {\em PLoS One}, 7(5):e36026, May 2012.

\bibitem[\protect\citeauthoryear{{Oxford Vaccine Group}}{2020}]{flu}
{Oxford Vaccine Group}.
\newblock Inactivated flu vaccine.
\newblock Available as
  \url{https://vk.ovg.ox.ac.uk/vk/inactivated-flu-vaccine}, 2020.

\bibitem[\protect\citeauthoryear{Padmanabhan \bgroup \em et al.\egroup
  }{2021}]{padmanabhan2021reinforcement}
Regina Padmanabhan, Nader Meskin, Tamer Khattab, Mujahed Shraim, and Mohammed
  Al-Hitmi.
\newblock Reinforcement learning-based decision support system for covid-19.
\newblock {\em Biomedical Signal Processing and Control}, 68:102676, 2021.

\bibitem[\protect\citeauthoryear{Perkins and España}{2020}]{Perkins}
T.~A. Perkins and G.~España.
\newblock {{O}ptimal {C}ontrol of the {C}{O}{V}{I}{D}-19 {P}andemic with
  {N}on-pharmaceutical {I}nterventions}.
\newblock {\em Bull Math Biol}, 82(9):118, 09 2020.

\bibitem[\protect\citeauthoryear{Raffin \bgroup \em et al.\egroup
  }{2021}]{JMLR}
Antonin Raffin, Ashley Hill, Adam Gleave, Anssi Kanervisto, Maximilian
  Ernestus, and Noah Dormann.
\newblock Stable-baselines3: Reliable reinforcement learning implementations.
\newblock {\em Journal of Machine Learning Research}, 22(268):1--8, 2021.

\bibitem[\protect\citeauthoryear{Rao}{2020}]{bellman}
Ashwin Rao.
\newblock Discrete versus continuous markov decision processes.
\newblock Available as
  \url{https://web.stanford.edu/class/cme241/lecture_slides/DiscreteVSContinuous.pdf},
  2020.

\bibitem[\protect\citeauthoryear{Ritto \bgroup \em et al.\egroup
  }{2021}]{uncertain}
Thiago~G Ritto, Americo Cunha~Jr, and David~A.W. Barton.
\newblock {Parameter calibration and uncertainty quantification in an SEIR-type
  COVID-19 model using approximate Bayesian computation}.
\newblock In {\em {3rd Pan American Congress on Computational Mechanics (PANACM
  2021)}}, Rio de Janeiro, Brazil, November 2021.

\bibitem[\protect\citeauthoryear{Schulman \bgroup \em et al.\egroup
  }{2017}]{ppo}
John Schulman, Filip Wolski, Prafulla Dhariwal, Alec Radford, and Oleg Klimov.
\newblock Proximal policy optimization algorithms.
\newblock {\em CoRR}, abs/1707.06347, 2017.

\bibitem[\protect\citeauthoryear{Song \bgroup \em et al.\egroup
  }{2020}]{song2020reinforced}
Sirui Song, Zefang Zong, Yong Li, Xue Liu, and Yang Yu.
\newblock Reinforced epidemic control: Saving both lives and economy.
\newblock {\em arXiv preprint arXiv:2008.01257}, 2020.

\bibitem[\protect\citeauthoryear{Tariq \bgroup \em et al.\egroup
  }{2021}]{epipolicy}
Zain Tariq, Miro Mannino, Mai Le~Xuan~Anh, Whitney Bagge, Azza Abouzied, and
  Dennis Shasha.
\newblock {\em Planning Epidemic Interventions with EpiPolicy}, page 894–909.
\newblock Association for Computing Machinery, New York, NY, USA, 2021.

\bibitem[\protect\citeauthoryear{Wang \bgroup \em et al.\egroup
  }{2019}]{benchmark}
Tingwu Wang, Xuchan Bao, Ignasi Clavera, Jerrick Hoang, Yeming Wen, Eric
  Langlois, Shunshi Zhang, Guodong Zhang, Pieter Abbeel, and Jimmy Ba.
\newblock Benchmarking model-based reinforcement learning.
\newblock {\em CoRR}, abs/1907.02057, 2019.

\bibitem[\protect\citeauthoryear{Zong and Luo}{2022}]{zong2022reinforcement}
Kai Zong and Cuicui Luo.
\newblock Reinforcement learning based framework for covid-19 resource
  allocation.
\newblock {\em Computers \& Industrial Engineering}, 167:107960, 2022.

\end{thebibliography}
